\documentclass[10pt,conference]{IEEEtran}
\usepackage{graphicx}
\usepackage{subcaption}
\usepackage{latexsym}
\usepackage{amsxtra} 
\usepackage{amsmath}
\usepackage{amsfonts}
\usepackage{marvosym}
\usepackage{scalefnt}
\usepackage{xcolor,tabularx}
\usepackage{makecell}
\usepackage{booktabs,multicol,multirow}
\usepackage{algorithm}
\usepackage{comment}
\usepackage{cite}
\usepackage{url}
\usepackage[noend]{algpseudocode}
\usepackage{fancyhdr}
\begin{document}

\title{Genetic Programming with Transformer-Based Mutation for Approximate Circuit Design}

\author{
\IEEEauthorblockN{Ondrej Galeta and Lukas Sekanina}
\IEEEauthorblockA{\scalefont{1.0}{Brno University of Technology, Faculty of Information Technology, Brno, Czech Republic}\\
xgalet06@stud.fit.vut.cz, sekanina@fit.vut.cz
}} 
\maketitle
\thispagestyle{fancy}

\begin{abstract}
A recent trend is to leverage machine learning models to improve the evolutionary design and optimization process. We propose a novel transformer-based mutation operator for Cartesian genetic programming (CGP) for the automated design of approximate arithmetic circuits. We introduce a hybrid scheme for CGP in which the proposed mutation operator is switched with the standard mutation operator to prevent stagnation of the circuit approximation process. We also develop a new training scheme for the underlying transformer that utilizes training vectors composed of thousands of CGP chromosomes representing various approximate multipliers. For several target error constraints, the approximate multipliers evolved with CGP utilizing the transformer-based mutation achieve better trade-offs than the highly optimized designs available in the state-of-the-art EvoApproxLib library of approximate circuits. Although both training and evolutionary processes are computationally demanding, they appear to be necessary steps for improving existing approximate circuits and producing new, potentially patentable circuit designs.
\end{abstract}


\section{Introduction}

\emph{Approximate computing} is a design paradigm that leverages an application’s tolerance to error in order to improve metrics such as latency, power consumption, and resource usage~\cite{Mittal:2016}. This approach is particularly important in both hardware and software implementations of deep neural networks (DNNs), where substantial energy savings can be achieved by employing simplified—and therefore approximate—data types, arithmetic units, memory subsystems, training, and inference procedures~\cite{ArmeniakosZSH23}. Because DNNs exhibit inherent error resilience~\cite{ArmeniakosZSH23}, carefully introduced approximations in their implementations typically result in only modest, and thus acceptable, degradation in accuracy.

Approximate implementations of arithmetic circuits, such as adders and multipliers, are obtained by simplifying their underlying logic functions. Approximation techniques range from manually designed methods such as bit-width reduction, removal of less significant circuit components, and logic resynthesis, to automated, algorithmic approaches~\cite{Jiang:axc:surv:2020}. One prominent method is \emph{Cartesian Genetic Programming} (CGP)~\cite{miller:cgp:book}, which was used to create the widely adopted EvoApproxLib library of approximate arithmetic circuits~\cite{mrazek:date:17, Mrazek:2020:approxMultipliersForCNN}. These circuits are publicly available\footnote{\url{https://ehw.fit.vutbr.cz/evoapproxlib/}} for various bit widths and are accompanied by comprehensive characterization data, including metrics such as area, latency, power consumption, and multiple error measures. 
They exhibit excellent trade-offs between the key design objectives and thus outperform other approximate arithmetic circuits under the target quality criteria~\cite{Mrazek:2020:approxMultipliersForCNN, Jiang:axc:surv:2020}.

In addition to a highly optimized fitness evaluation that enables fast and precise error calculation even for complex circuits (e.g., 32-bit approximate multipliers~\cite{ceska:iccad17}), the CGP-based approximation method used to generate EvoApproxLib relies on standard Cartesian Genetic Programming with a simple point mutation operator (more advanced strategies and mutation operators were proposed later in~\cite{Ceska:sagtree:2022}). The circuits included in the final EvoApproxLib collection were selected from numerous computationally expensive CGP runs~\cite{Mrazek:2020:approxMultipliersForCNN}. Consequently, further improving their properties is both challenging and computationally expensive.

A recent trend is to leverage \emph{machine learning} (ML) models to improve the evolutionary process~\cite{Wu:LLM:EA:2025, Hemberg2025, Sobania:tec:2025}.
In this paper, we propose a novel \emph{transformer-based mutation operator} that, when integrated into standard CGP, enhances the search process. As far as we know, this is the first approach combining transformers with CGP. In some cases, the approximate multipliers evolved with CGP utilizing the proposed transformer-based mutation achieve better trade-offs than the highly optimized designs available in EvoApproxLib. Our mutation operator is trained on a dataset comprising thousands of approximate multipliers. Although both transformer training and the evolutionary design process are computationally demanding, they appear to be necessary technical steps in improving existing approximate circuits and producing the new, potentially patentable circuit designs.

The contributions of this paper are as follows: 
\begin{itemize}
\item We develop a new transformer-based mutation operator and a hybrid scheme for CGP, in which the proposed mutation is alternated with the standard mutation to prevent stagnation in the circuit approximation process. 
\item We propose a training scheme for the underlying transformer that utilizes thousands of training vectors composed of CGP chromosomes representing various approximate multipliers.
\item For 8-bit approximate multipliers constrained on the worst-case error (WCE), we show that CGP utilizing the transformer-based mutation can create circuits showing better trade-offs between WCE and area than the highly optimized circuits available in the state-of-the-art EvoApproxLib library of approximate circuits.
\end{itemize}

The remainder of the paper is organized as follows.
Section~\ref{sec:soa} reviews the principles of approximate arithmetic circuit design, with an emphasis on CGP-based approximation methods and the application of machine learning in genetic programming.
Section~\ref{sec:method} introduces the proposed transformer-based mutation operator for CGP, including the transformer training procedure.
Section~\ref{sec:setup} summarizes the experimental setup for both transformer training and the CGP-based circuit approximation framework.
Section~\ref{sec:results} presents the experimental results, including a comparison of CGP performance using the standard and the proposed mutation operators.
Finally, Section~\ref{sec:concl} concludes the paper.

\section{Related Work}
\label{sec:soa}

\subsection{Approximate Circuits}

As this paper deals with functional approximation of arithmetic circuits, in particular multipliers, we will not discuss other areas of approximate computing techniques that are summarized in~\cite{ACsurvey:ACM:2020}.
According to a survey~\cite{Jiang:axc:surv:2020}, approximate implementations of arithmetic circuits can be obtained at three main levels: 
At the \emph{algorithm level}, the exact standard multiplication is approximated by another algorithm that produces inexact but sufficiently good results (e.g., a logarithm-based multiplier (ALM)~\cite{ALM}).
At the \emph{architecture level}, a suitable exact circuit implementation is selected and then modified at different stages (partial product generation, accumulation, compressors, etc.) according to a human-created strategy or heuristically. A typical example is a broken array multiplier (BAM)~\cite{Mahdiani:TCSI2009} in which certain columns and rows of the partial product array are removed. 
At the \emph{circuit level}, gate- and transistor-level approximation techniques such as circuit rewriting~\cite{mrazek:date:17, WitschenMAP19, Yi:GPTAC:2024} are applied to create simplified circuits.

The design objective is to create approximate circuits that show excellent trade-offs between hardware properties and error. The error of an approximate arithmetic circuit can be quantified using several metrics, such as the \emph{worst-case error} (WCE) defined in Eq.~\ref{eq:WCE} and the \emph{mean absolute error} (MAE) defined in Eq.~\ref{eq:MAE}.
\begin{equation}
\text{WCE} = \max_{\forall i} \left| O_{\text{approx}}^{(i)} - O_{\text{orig}}^{(i)} \right|
\label{eq:WCE}
\end{equation}
\begin{equation}
\mathrm{MAE} = \frac{\sum_{\forall i} \left| O_{\text{approx}}^{(i)} - O_{\text{orig}}^{(i)} \right|}{2^{n}}
\label{eq:MAE}
\end{equation}
where $O_{\text{orig}}^{(i)}$ is the original output and $O_{\text{approx}}^{(i)}$ is the approximate output (in the decimal system) for a given input $i$ of an $n$-bit circuit (i.e., $n=2\cdot k$ for a $k$-bit multiplier). Hardware properties such as area, latency, and power consumption are typically estimated during the approximation process to reduce its execution time. The resulting approximate circuits are fully characterized using professional electronic circuit design tools.

\subsection{Cartesian Genetic Programming}
\label{sec:cgp}

One of the established circuit approximation methods is Cartesian Genetic Programming (CGP)~\cite{miller:cgp:book}. In this section, we briefly review the standard CGP representation; the specific modifications introduced for the proposed approach are described in Section~\ref{sec:method}.

In CGP, a candidate gate-level circuit with $n_i$ primary inputs and $n_o$ outputs is represented as a two-dimensional grid of computational nodes arranged in $n_r$ rows and $n_c$ columns. Each primary input and each node is assigned a unique identifier (ID). The circuit structure is encoded as a vector of integer values, known as the chromosome.
Each node typically represents a two-input logic gate and is described by three integers in the chromosome. The first two integers specify the IDs of the nodes connected to the gate inputs, while the third integer selects the logic function from a set of functions $\Gamma$ available to CGP. The primary outputs are defined by an additional vector of $n_o$ integers, each determining the ID of the node driving the corresponding output. The size of the chromosome that represents a circuit composed of two input gates is $3\cdot n_r \cdot n_c +n_o$.

\begin{figure}[h] 
    \centering
    \includegraphics[width=0.45\textwidth]{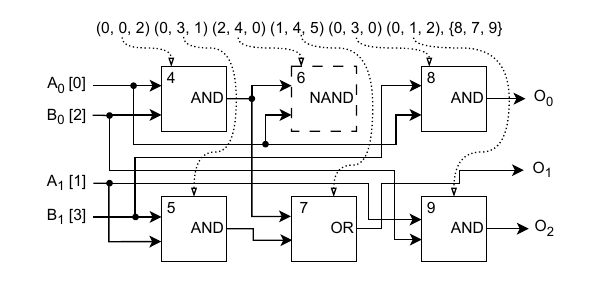}  %
    \caption{Example graph and chromosome representation of a combinational circuit in CGP with the parameters $n_i=4, n_o=3, n_c=3,n_r=2,\Gamma=\{0^{AND},1^{OR},2^{NAND}\}$. Every node is represented by its operation and two inputs; the nodes are indexed top to bottom, left to right.}
    \label{fig:cgp}
\end{figure}

CGP usually relies on mutation as its genetic operator. Mutation may alter the logic function of a node, modify the connections of its inputs, or change the assignment of primary outputs. Feedback connections are not permitted, ensuring a feed-forward circuit structure. Figure~\ref{fig:cgp} illustrates a CGP representation of a two-input, three-output circuit along with its corresponding encoding. It should be noted that not all nodes in the CGP representation necessarily contribute to the final circuit structure, as some nodes may remain inactive (see dashed blocks in Fig.~\ref{fig:cgp}). The fitness function can be designed to simultaneously promote functional correctness with respect to the target specification and to penalize the use of unnecessary gates, thereby encouraging more compact circuit implementations. The search process in CGP typically follows a $(1+\lambda)$ evolutionary strategy, where $\lambda$ offspring individuals are generated from the parent through mutation, and the best individual among the parent and offspring is selected as the parent for the next generation~\cite{miller:cgp:book}.

CGP has been employed for the design of approximate arithmetic circuits in EvoApproxLib~\cite{mrazek:date:17}. As described in~\cite{mrazek:date:17, ceska:iccad17}, the process begins with a user-selected exact gate-level circuit, which is subsequently modified through CGP-based mutations to obtain circuits exhibiting desired properties.
While Mrazek et al. utilized the NSGA-II multi-objective evolutionary algorithm to construct EvoApproxLib, Češka et al.~\cite{ceska:iccad17} focused on minimizing the relative circuit area—closely correlated with power consumption—subject to a user-defined constraint on the worst-case error. Various alternative search strategies for approximate circuit design were explored in the literature, e.g., \cite{WitschenMAP19}.

\subsection{Transformers in Circuit Design}

The concept of transformer neural networks with self-attention was first introduced in ~\cite{vaswani:2017:attentionneed} and became widely popular in natural language processing following the introduction of BERT~\cite{devlin:2019:bert}. BERT is an \emph{encoder-only transformer} that leverages bidirectional self-attention to produce rich contextual embeddings of tokens. This approach allows the model to capture complex patterns in a given sequence. Inspired by this success, recent works, outside the evolutionary computation field, have explored adapting transformer models to enhance or perform the automated design in various domains, including digital circuit design
~\cite{VeriGen:2024, Yi:GPTAC:2024}. For example, Yi et al.~\cite{Yi:GPTAC:2024} used a generative pre-trained model that automatically generates approximate digital circuits from a tokenized hardware description representation. Using this approach, they designed approximate multipliers with configurable accuracy. While the aforementioned articles approach transformers from different perspectives, both demonstrate the transformer’s ability to learn patterns from data in a given design problem and propose contextually meaningful modifications or whole solutions.

\subsection{Evolutionary Machine Learning}

The \emph{evolutionary machine learning} deals with the interaction between machine learning (ML) and evolutionary algorithms (EA). According to~\cite{EMLbook:2023}, three major areas are observed:  the use of EA to improve ML methods; the use of ML techniques to improve EA; and the application of EA to problems traditionally solved by standard ML approaches. With growing interest in large language models (LLMs), the interaction between LLMs and EAs was addressed in surveys~\cite{Wu:LLM:EA:2025, Hemberg2025, Sobania:tec:2025}. For example, Hemberg et al.~\cite{Hemberg2025} categorized the interaction between genetic programming (GP) and LLM as follows: (i) LLM as part of GP (e.g., LLM is a smart mutation operator~\cite{shem:2025:evoTrans}); (ii) LLM supporting GP (e.g., test case generation by LLM, representation design~\cite{CaetanoTP23}); (iii) GP supporting LLM (e.g., prompt optimization).  

For example, Shem-Tov et al.~\cite{shem:2025:evoTrans} used the BERT Transformer as a context-aware mutation operator in GP to intelligently suggest node replacements within program trees based on the tree structure and historical fitness data. They show that their transformer-based mutation operators can improve convergence. Teixeira and Pappa~\cite{TexPappa:gecco25} developed a transformer-encoder as a surrogate to evaluate pairs of solutions and determine which one is better/worse than the other. The trained encoder was compared against a traditional GP that evaluated fitness at each generation.

\section{Proposed Method}
\label{sec:method}

Before we introduce the proposed evolutionary approximation method utilizing CGP with a transformer-based mutation operator, we first describe the standard CGP as used for circuit approximation in~\cite{mrazek:date:17,  Mrazek:2020:approxMultipliersForCNN}. This will constitute our baseline for all experiments. Then, the proposed transformer, training strategy, and transformer's incorporation in CGP will be presented.

\subsection{Standard CGP for Circuit Approximation}
The CGP configuration used in this work follows the description provided in Section~\ref{sec:cgp} and the implementation reported in~\cite{mrazek:date:17, Mrazek:2020:approxMultipliersForCNN}. The nodes are arranged in a single row (i.e., $n_r = 1$). Each node represents a logic gate and may receive inputs either from the primary inputs or from preceding nodes in the sequence.
Each gate is associated with a value representing its estimated area on a chip. The total circuit area is computed as the sum of the areas of all active gates in the evolved circuit.

Each chromosome is evaluated using the fitness function defined in Eq.~\ref{eq:fitness}, where WCE$(\widetilde{M})$ denotes the worst-case error of a candidate multiplier $\widetilde{M}$ computed over all possible input vectors. The optimization objective is to minimize the circuit \emph{cost} (i.e., area) subject to two constraints: (1) WCE$(\widetilde{M}) \leq \epsilon$, where $\epsilon$ is a user-defined error threshold, and (2) all multiplications involving zero must be exact (i.e., $\mathrm{WCE}_{zr}(\widetilde{M}) = 0$), a requirement that is critical for deploying approximate multipliers in deep neural networks~\cite{Mrazek:2020:approxMultipliersForCNN}.
\begin{equation}
F(\widetilde{M}, \epsilon) = 
\begin{cases} 
cost(\widetilde{M}) & \text{if WCE}(\widetilde{M}) \leq \epsilon  ~~\wedge \\
& \text{WCE}_{zr}(\widetilde{M}) = 0\\
\infty & \text{otherwise.}
\end{cases}
\label{eq:fitness}
\end{equation}

Exhaustive circuit simulation is performed with an optimized implementation in C that leverages the Single Instruction Multiple Data (SIMD) paradigm and considers only active nodes (i.e., nodes that affect the output of the represented circuit), following ideas similar to those explored in~\cite{Vasicek:2012:efficientCGP,Hrbacek:2014:efficientCGP}. Evaluation is carried out in stages. First, the circuit size is checked against the parent circuit size, then a zero-multiplication test is applied, and finally, the remaining test vectors are evaluated to identify circuits that outperform the current best. This staged evaluation allows circuits to be rejected quickly, so that not every chromosome undergoes all evaluation steps.

The evolutionary process is initialized with a chromosome corresponding to a known accurate multiplier. In each generation, if no offspring exhibits strictly better fitness than the parent, an offspring with fitness equal to that of the parent is selected as the new parent. If all offspring have worse fitness, the parent is retained unchanged. This selection strategy exactly follows Miller's original approach in CGP~\cite{miller:cgp:book}.

New offspring are generated exclusively through single-point mutation, where exactly one mutation is applied to an active node in the chromosome per generation. 
In the standard, non-transformer setting, mutation alters a node by randomly modifying either its predefined function or one of its two input connections, selecting an alternative from the set of permissible options with uniform probability.
The evolutionary process terminates after a predefined number of generations or once the allocated computational time has been exhausted.

\subsection{Transformer-based Mutation}

\begin{figure}[h] 
    \centering
    \includegraphics[width=
    \columnwidth]{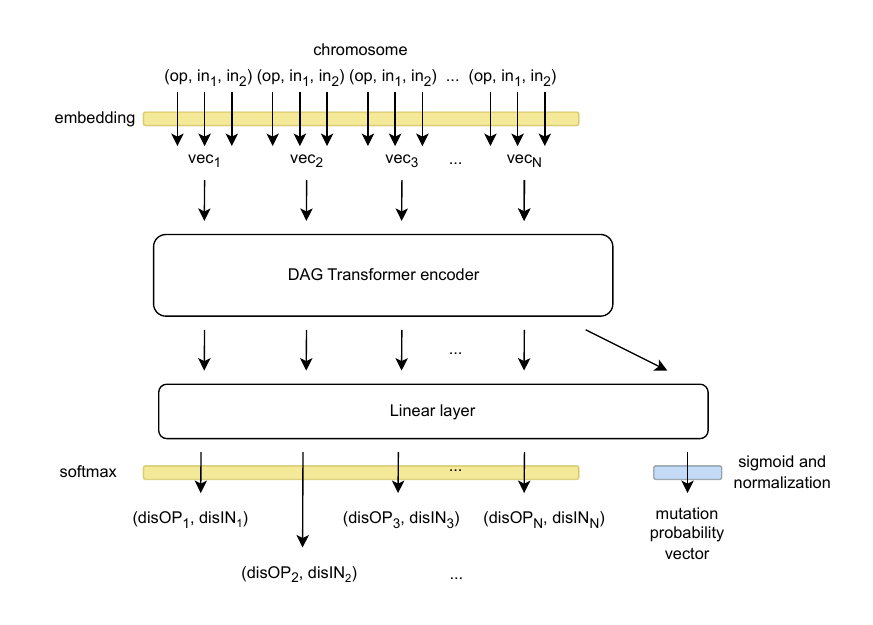}  %
    \caption{Proposed transformer-based mutation operator}
    \label{fig:transformer_proposed}
\end{figure}

The operation of the proposed transformer-based mutation is visualized in Fig.~\ref{fig:transformer_proposed}. The transformer takes as input a chromosome and outputs probability distributions that guide both the location and the type of mutation applied to individual nodes. 
During mutation, a \emph{node position} is sampled from a mutation probability vector (the blue part of Fig.~\ref{fig:transformer_proposed}). Then, one of the determined node's inputs or its function is selected uniformly at random to be replaced, and the replacement is sampled from a probability distribution specific to the given node position (the yellow output vector in Fig.~\ref{fig:transformer_proposed}).
It should be emphasized that, unlike the standard CGP, we omit the $n_o$ integers of the chromosome that describe the connection of primary outputs. In our model, the primary outputs are directly taken from the last $n_o$ gates. The model's functionality is similar to that of an autoencoder. The core of the model is a BERT-style encoder-only transformer, as described in~\cite{devlin:2019:bert}, with several small refinements. The model encodes each node using a 64-dimensional embedding.

Each node is represented as an embedding by independently encoding its three components and concatenating the results into a single vector. This embedding is then adjusted (biased) by incorporating embeddings of all preceding nodes to which the node is connected if such connections exist.
The bias is made by adding embedded vectors of preceding nodes to the node embedding, as shown in Eq.~\ref{eq:embd}, where $e$ is the embedding and $c_{\mathrm{par}}$ is the coefficient determining the emphasis on preceding nodes.

\begin{equation}
e_i' = e_i + c_{\mathrm{par}} \cdot 
\frac{\sum_{j \in \text{valid parents of }i} e_{j}}{\#\text{valid parents of }i}
\label{eq:embd}
\end{equation}

The embedding of a node is therefore no longer determined only by the node itself, but also by the embeddings of preceding nodes in the circuit graph. This modification allows the transformer to explicitly prioritize preceding nodes as the most influential context for each node while still capturing global information about the entire circuit.

\subsection{Training Data and Transformer Training}

\begin{figure}[ht]
\centering
\includegraphics[width=\columnwidth]{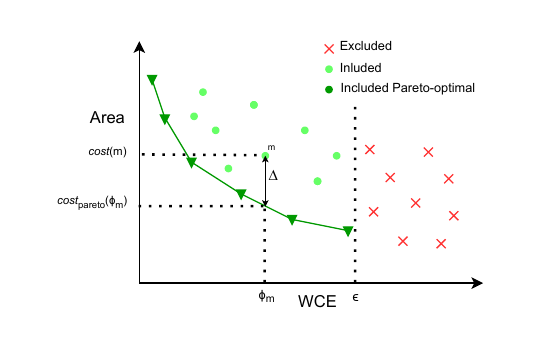}
\caption{The principle of multiplier selection from a dataset of approximate multipliers for transformer model training.}
\label{fig:filtering}
\end{figure}

\begin{figure}[ht]
\centering
\includegraphics[width=0.99\columnwidth]{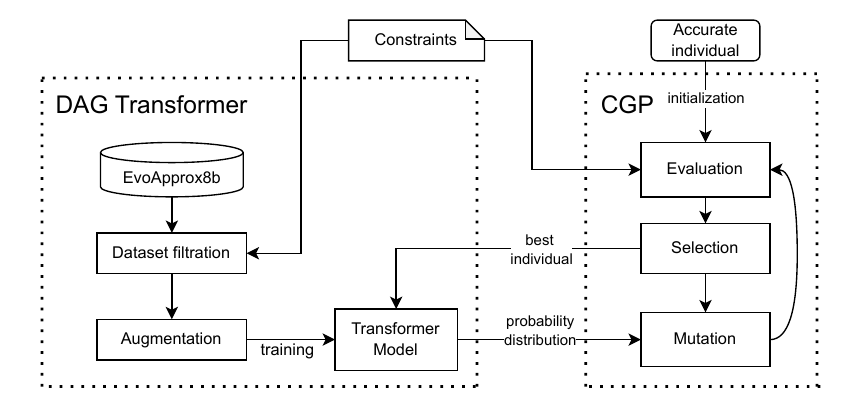}
\caption{Transformer training and incorporating the trained transformer to the CGP loop.}\label{fig:dataflowoverview}
\end{figure}

As this paper deals with 8-bit approximate multipliers, we extracted only these circuits from EvoApproxLib. The initial dataset $D$, which is available for training the proposed transformer, contains 39,699 unsigned 8-bit approximate multipliers. For each target WCE, a specialized transformer model is created and trained using a subset of $D$, denoted $D_\text{filtered}$. In this paper, only two objectives are considered for the approximate multiplier design: WCE and area. The dataset and transformer training preparation steps are as follows (see also Fig.~\ref{fig:dataflowoverview}):
\begin{itemize}
    \item The input of the method is (i) a constraint $\epsilon$ that determines the maximal acceptable WCE and (ii) a data set of approximate multipliers $D$.
    
    \item In the data set $D_{valid}$, only valid approximate multipliers w.r.t. WCE are kept, i.e. $\forall m$, $m \in D_\text{valid}: m \in D \wedge \text{WCE}(m) \leq \epsilon \wedge \text{WCE}_\text{zr} = 0$.
    
    \item A Pareto front reflecting the WCE and the multiplier area is constructed from circuits of $D_{valid}$. The Pareto front is interpolated using an auxiliary curve (depicted in green in Fig.~\ref{fig:filtering}) to build ideal (golden) solutions needed for training. Let $cost_\text{pareto}(\phi)$ be the area of a hypothetical circuit multiplier with $\text{WCE}=\phi$ lying on this curve, as seen in Fig.~\ref{fig:filtering}. 
    
    \item We then calculate a weight $a$ that determines ``attractiveness" of a given multiplier $m$; $\forall m\in D_\text{filtered}$, $a=e^{-d\Delta}$, where $d$ is the decay coefficient, representing how strictly we want to rely on multipliers that are closer in area to the interpolated curve. The symbol  $\Delta$, defined as $\Delta=cost(m) - cost_\text{pareto}(\phi_m)$, represents the size difference of multiplier $m$ and hypothetical circuit multiplier with $\text{WCE}=\phi_m$ lying on the interpolated curve, where $\phi_m$ is the WCE of multiplier $m$. The parameter $d$ was empirically validated as $d=0.01$. 
    
    \item For each active node $n$ in multiplier $m$ with WCE $= \phi$, a random mutation is performed and a new value of WCE, denoted $e_i$, is calculated for the multiplier created by that mutation.  
    This step is repeated $L$ times for each active node in the chromosome. The ``sensitivity" approximation, $\tilde{S}_n = \log\left( \left( \prod_{i=1}^{L} \frac{e_i}{e_0} \right)^{\frac{1}{L}} \right)$, is determined. The sensitivity across all nodes of the chromosome is then normalized to $0 < \tilde{S}_n \leq 1$. Higher $L$ gives better $\tilde{S}$ approximation but also significantly raises computational demands. We found that $L=8$ is sufficient for our case. 

    \item Each time a multiplier is taken from the dataset $D_\text{filtered}$, it is augmented, i.e., its new versions are created to expand and improve the dataset. \emph{Augmentation} in our case involves shuffling the positions of nodes in a grid in a way that the behavior of the represented multiplier remains unchanged. It helps the transformer to ``understand'' the syntax of CGP chromosomes. 
    
\end{itemize}

 The transformer model is trained using a \emph{ masking-based learning objective}. Randomly chosen elements of the input chromosome are masked, and the partially observed chromosome is provided as the input to the model. The model's output then represents the distribution for each node input and gate function, as well as the sensitivity value for each node. Despite each gate having two inputs, the input distribution is only one, and it is shared between them. The loss is computed as follows:

\begin{itemize}

\item The \emph{loss for the gate function} $\mathcal{L}_{\text{op}}$ and \emph{the gate inputs} $\mathcal{L}_{\text{input}}$ are calculated using the binary cross-entropy loss between the predicted and the true model output. 

\item The \emph{sensitivity loss} $\mathcal{L}_{\text{sens}}$ is computed as the MSE between the true and predicted values.

\item %
To reduce model hallucinations, when a high probability is assigned to functions or inputs that do not match the true values, a \emph{confidence loss} is introduced, following the approach proposed in~\cite{Pereyra:2017:Regularizing},
\begin{equation}
\mathcal{P}_{\text{conf}}
=\sum_{h=1}^{H}
\sum_{c=1}^{C}
\left(\sigma\left(o_{h,c}\right)\cdot(1-y_{h,c})\right)^2,
\label{eq:confidenceLoss}
\end{equation} 
where $o_{h,c}$ represents the logit corresponding to node $l$ and class $c$, $C$ is the number of logits in the given category (inputs or gate function), $y_{h,c}$ is the target value, $\sigma(\cdot)$ denotes the sigmoid activation function, and $H$ 
is the number of all active nodes in the chromosome.
\item 
The \emph{total loss} is then computed as:
\begin{equation}
\mathcal{L}_{\text{total}} =
\hat{a}(\mathcal{L}_{\text{op}} + \mathcal{L}_{\text{input}} +
c_{\text{op}} \mathcal{P}_{\text{conf}}^{\text{op}} +
c_{\text{in}} \mathcal{P}_{\text{conf}}^{\text{in}}) +
c_{\text{sens}} \mathcal{L}_{\text{sens}}
\label{eq:total_loss}
\end{equation}
where $\hat{a}$ is the normalized attractiveness of $a$ within the batch and calculated during the preparation of the data set,
$\mathcal{P}_{\text{conf}}^{\text{op}}$ and $\mathcal{P}_{\text{conf}}^{\text{input}}$ are separated confidence losses for gate function and inputs, and $c$ are the weighting coefficients. 
\end{itemize}

\subsection{CGP with a Transformer-Based Mutation}

To make the transformer operator directly comparable with classical CGP mutation, the aim is to run the whole proposed CGP algorithm exclusively on the CPU. To reduce computational overhead caused by model inference, the mutation distribution is recomputed only when the best fitness value changes. Otherwise, it uses the same distribution as in the previous generation. The reason is that a mutation is applied only to one node element across the entire chromosome. This means that if the fitness is the same, the input chromosome is the same, or it provides a neutral mutation that has no effect on fitness.   

Another optimization stems from the idea that, in practice, our transformer model 
can produce useless results
due to a limited dataset. Sometimes, its output
can lead to a dead end, misleading mutations in a way that makes it very difficult to achieve progress in improving fitness. %
Hence, we propose a hybrid scheme that does not rely on the transformer model itself and still uses the standard CGP mutation operator as a backup, which is deployed when a long sequence of generations is carried out without any progress. The classical CGP mutation operator is then used until fitness improves again and the model distribution changes. This concept stems from the fact that the presented transformer model aims to indirectly model the fitness function. However, it is not capable of modeling it perfectly, which can guide the search method to less promising regions of the search space. 
Switching to the classical CGP mutation operator can then be understood as a way to escape from these regions more easily. 

The general scheme of the proposed method is shown in Fig.~\ref{fig:dataflowoverview} and a high-level pseudocode of CGP with a transformer operator is given in Algorithm~\ref{algo:cgp_transformer_pseudocode}. Starting from an initial best individual, the algorithm iteratively generates offspring through mutation over a fixed number of generations. When stagnation is not observed for a given maximal number of generations, mutations are guided by a probability distribution given by the proposed transformer-based \textit{model}($\cdot$) learned from the current best individual, biasing the search toward promising regions of the design space. If stagnation persists beyond a predefined limit, the algorithm switches to standard uniform random mutation to restore exploration. The best individual is updated in each generation based on fitness evaluation, and the process continues until the termination condition is met, returning the best evolved solution.

\begin{algorithm}
\caption{CGP with transformer}
\label{algo:cgp_transformer_pseudocode}
\begin{algorithmic}[1]
\Require Initial individual $bestIndv$, number of generations $gen$, 
stagnation limit $stagMax$, population size $\lambda$
\Ensure Best evolved individual $bestIndv$
\Procedure{CGP\_TRANS}{$bestIndv, gen, stagMax, \lambda$}
    \State $mDistr \gets model(bestIndv)$
    \State $stagCnt \gets 0$
    \For{gen}
        \If{$stagCnt < stagMax$}
            \State $pop \gets mutate\_trans(bestIndv,\lambda, mDistr)$
        \Else
            \State $pop \gets mutate\_uni(bestIndv, \lambda)$
        \EndIf
         \State $prevBestFit \gets fitness(bestIndv)$
         \State $bestIndv \gets select\_best(pop)$
        \If{$prevBestFit > fitness(bestIndv)$}
            \State $stagCnt \gets 0$
            \State $mDistr \gets model(bestIndv)$
        \Else
            \State $stagCnt \gets stagCnt + 1$
        \EndIf
    \EndFor
    \State \Return $bestIndv$
\EndProcedure
\end{algorithmic}
\end{algorithm}

\section{Setup}
\label{sec:setup}

\subsection{Transformer and Its Training}

The transformer model employs 4 attention heads, 6 transformer layers, and a feed-forward network with 256 hidden units. The embedding bias introduced in Eq.~\ref{eq:embd} is $c_{par}=0.2$. These parameters were experimentally proven to provide a good trade-off between computational complexity with respect to the nature of computational resources and the information gain this model can provide. %
The larger transformer models tend to provide only a very small increase in the quality of the distributions provided to the proposed operator, with an exponential increase in time required for a single inference. Smaller models struggle to capture complex data dependencies.
The masking ratio is linearly scheduled from $5\%$ to $30\%$ across epochs (curriculum learning).
Three different transformer models are initially trained for WCE thresholds $\epsilon = 3\%$, $4\%$, and $5\%$ with the corresponding datasets ($D_\text{filtered}$) containing 6551, 6800, and 10402 approximate multipliers, respectively. Other models were incorporated in Section~\ref{sec:comp:evoapproxlib}.
Training is held for 20 epochs with a batch size of 128. The loss function works best with the weights $c_{op}=0.1, c_{in}=0.2,$ and $c_{sens}=1$.

\subsection{CGP}

The CGP uses the following setting of parameters: $n_i = 2 \cdot 8 = 16, n_o = 16, n_r = 1, n_c = 600$, which were experimentally determined to provide a balance between sufficient space to obtain desired solutions and avoid unnecessary chromosome length.
The set of gates and their sizes are listed in Table~\ref{tab:gates}. These sizes correspond to the 45nm technology and are identical to those reported in~\cite{mrazek:date:17, ceska:iccad17, Mrazek:2020:approxMultipliersForCNN}.
Inspired in \cite{Mrazek:2020:approxMultipliersForCNN}, the number of offspring is $\lambda=4$,  
the maximum number of generations is $10^5$,
although it was rarely reached, as results are reported based on execution time rather than the number of generations. The CGP runs are uniformly seeded with six different accurate 8-bit multipliers (a ripple-carry array multiplier, two carry-save array multipliers, and three Wallace tree architectures). 

A single CGP run requires approx. 8 minutes on a \mbox{AMD EPYC 9124 16-Core Processor}, and one run is always assigned per core. 

\begin{table}[]
\centering
\caption{Sizes of the gates in $\mu m^2$ corresponding to the 45~nm technology.}
\label{tab:gates}
\begin{tabular}{|l|l|l|l|l|l|l|l|}
\hline
\textbf{Gate} & INV & AND & OR & XOR & NAND & NOR & XNOR \\ \hline
\textbf{Size} & 1.40 & 2.34 & 2.34 & 4.69 & 1.87 & 2.34 & 4.69 \\ \hline
\end{tabular}
\end{table} 

\section{Results}
\label{sec:results}

\begin{figure}[ht]
\centering
\includegraphics[width=0.9\columnwidth]{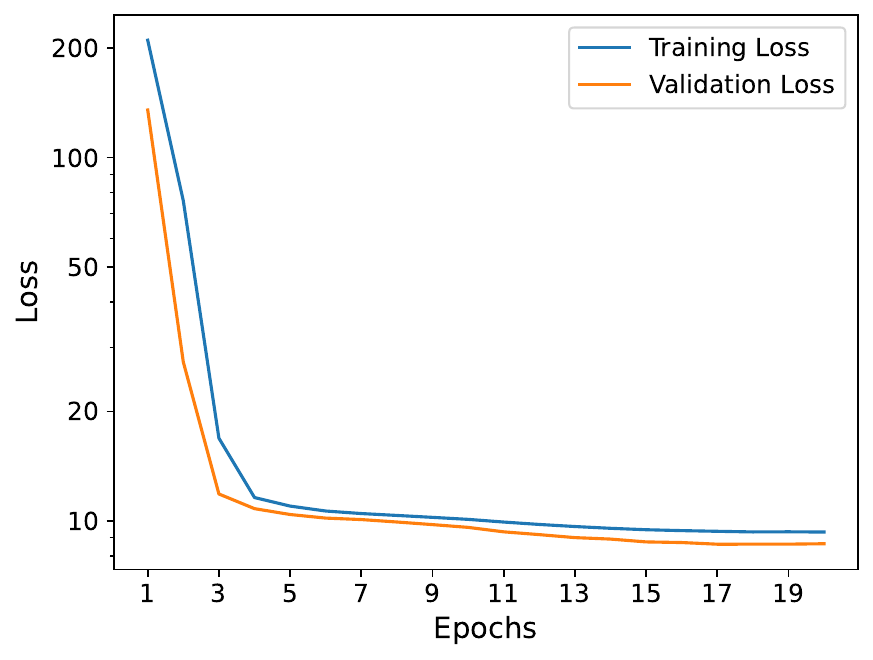}
\caption{Training progress of the proposed transformer model for $\epsilon=3\%$. }
\label{fig:train_progress}
\end{figure}

The proposed transformer model was trained using the loss function defined in Eq.~\ref{eq:total_loss} and the configuration described in Section~\ref{sec:setup}. As illustrated in Fig.~\ref{fig:train_progress}, the training process converges after approximately five epochs, with only marginal improvements in the loss thereafter (the model shown corresponds to $\epsilon = 3\%$). Training a single model requires approximately one hour on an NVIDIA A5000 GPU.

A total of 576 CGP independent runs are performed separately for each target WCE (i.e., for each transformer model), comparing the impact of both the standard and transformer-based mutation operators of CGP on the area of evolved multipliers. 
This high number of runs is crucial for obtaining statistically significant results, as CGP runs are stochastic and the transformer serves as a proxy model trained on relatively small datasets.

\subsection{Evaluation of the proposed mutation operator}

\begin{figure}[ht]
\centering
\includegraphics[width=0.99\columnwidth]{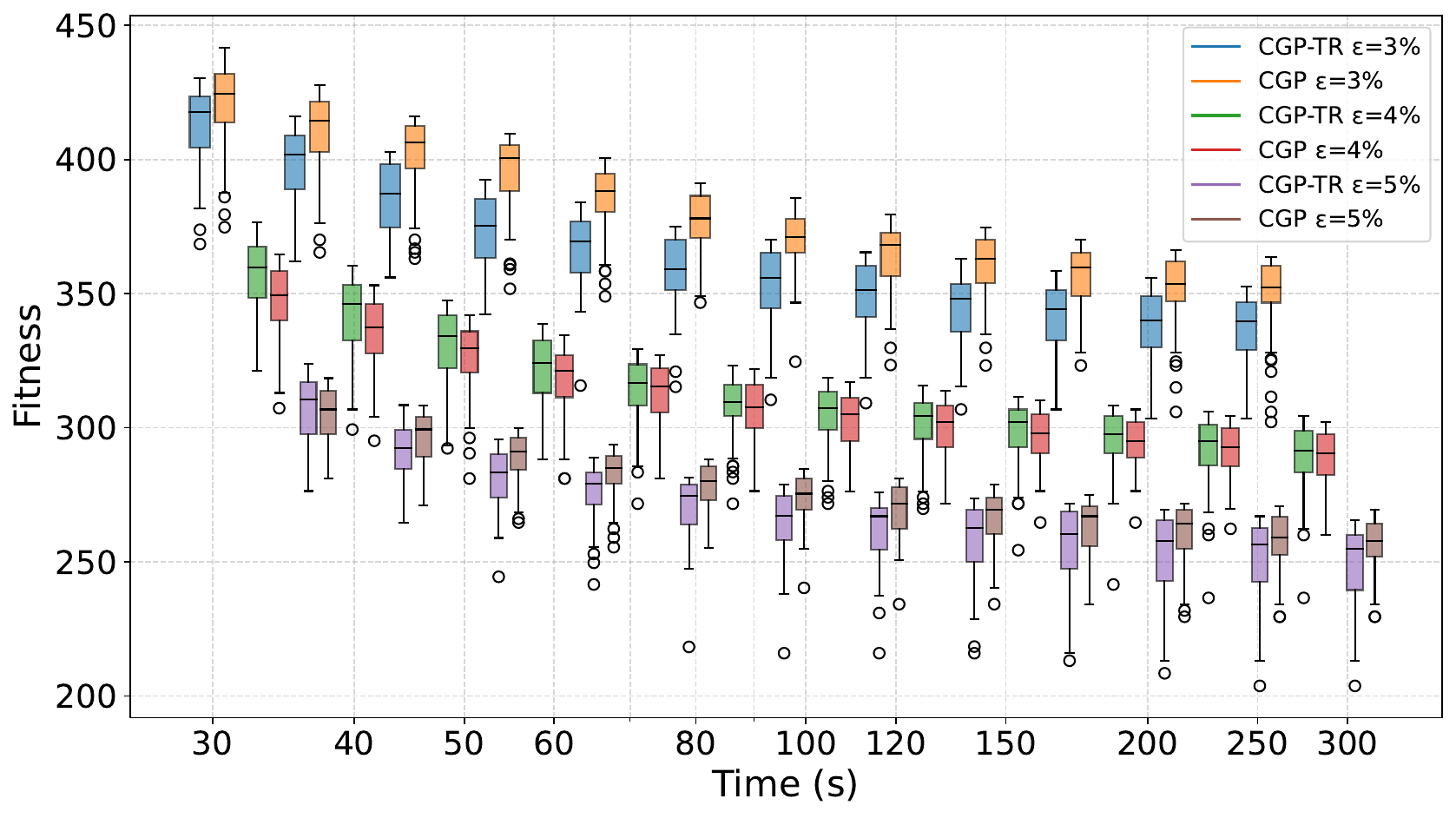}
\caption{Fitness progress in CGP utilizing standard mutation (CGP) and CGP with the transformer-based mutation operator (CGP TR) for three different WCE thresholds $\epsilon$. }
\label{fig:comparison:boxplot} 
\end{figure}

Fig.~\ref{fig:comparison:boxplot} compares the progress of the fitness of CGP utilizing the standard mutation and CGP with the hybrid transformer-based mutation (denoted CGP TR) for three target WCE $(\epsilon)$ levels. Each box plot shows only the top $10\%$ of multipliers at selected time points (identical for both algorithms) because we are primarily interested in obtaining approximate multipliers with excellent properties, i.e., those that improve upon state-of-the-art implementations. It can be observed that the fitness distribution of CGP TR is shifted toward lower (better) fitness values, as the objective is to minimize circuit size. 
The proposed model is slightly slower during the initial few seconds for $\epsilon=4$ and $\epsilon=5$, as improvements in the fitness function occur more frequently, requiring more frequent model inference to obtain updated output distribution. However, at that time, the fitness value for both algorithms is more likely to be poor. To present a fair comparison, the horizontal axis of Fig.~\ref{fig:comparison:boxplot} shows the execution time rather than the generation count. The standard CGP requires more generations than the CGP TR to produce the same quality result. 

To assess whether CGP TR provides statistically better results than standard CGP, 
we evaluated the performance of CGP and CGP TR using one-sided Mann–Whitney U tests on percentile-filtered samples, comparing distributions at $t = 150 \text{s}$ (approximately mid-run) and $t = 300 \text{s}$ (close to end of run).
For $\epsilon=3\%$, CGP TR significantly outperformed CGP at both time points ($p \approx 10^{-9} –  10^{-8}$), indicating a stable advantage throughout the run.
For $\epsilon=4\%$, no statistically significant differences were observed at either mid-run or end-run. 
For $\epsilon=5\%$, CGP TR showed a significant advantage at mid-run ($t = 150\text{s}, p = 0.000772$), and at the end of the run ($t = 300 \text{s}, p = 0.01813$).

\subsection{Setting of the Stagnation Limit}
\begin{figure}[ht]
\centering
\includegraphics[width=0.99\columnwidth]{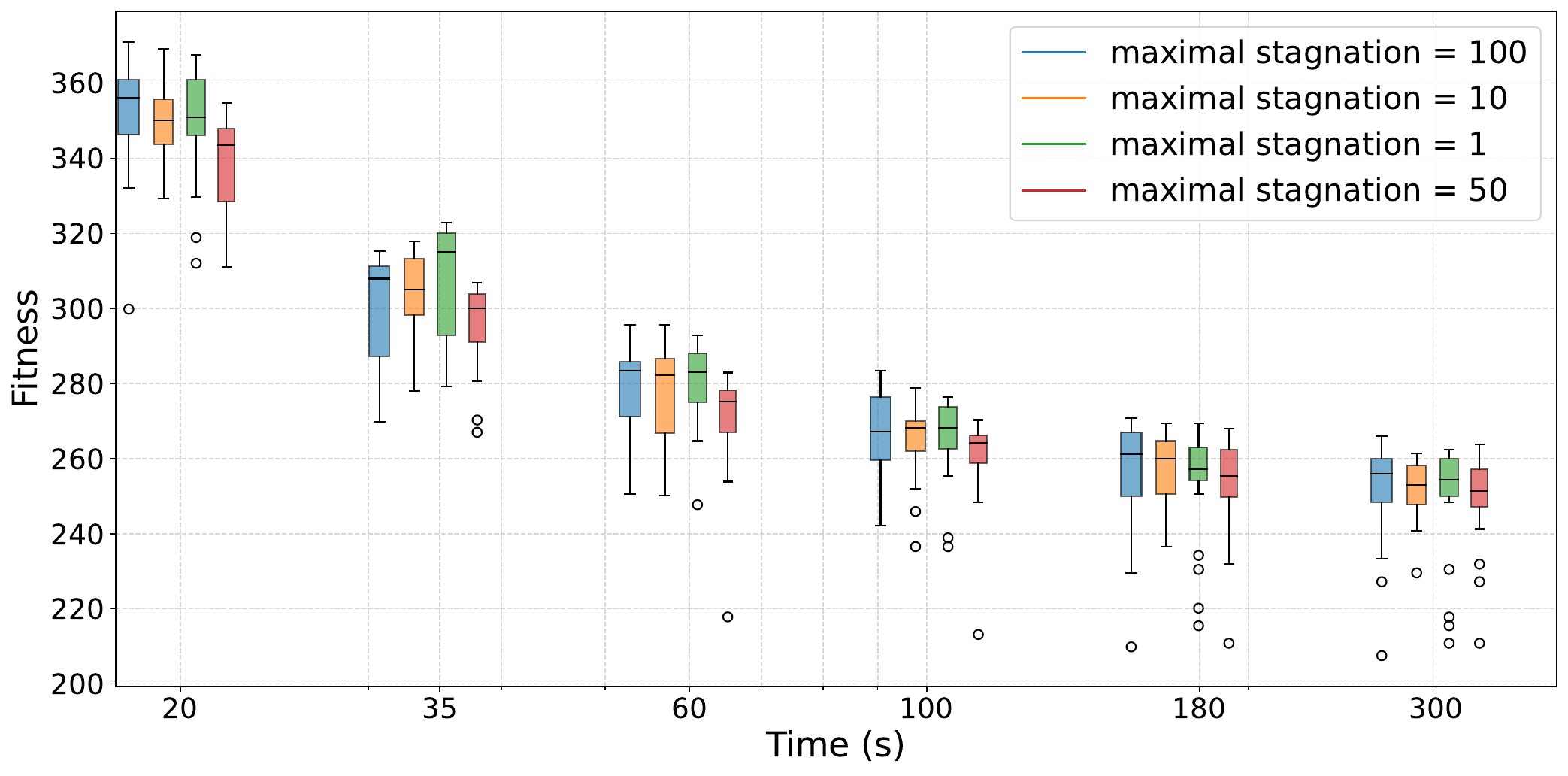}
\caption{Progress of CGP TR for different values of the maximal stagnation limit. } 
\label{fig:comparison:stagnationparameter}
\end{figure}

For CGP TR, a default maximum of 50 stagnating generations (\emph{stagMax} parameter of Alg. 1) is allowed before reverting to the standard mutation operator to recover progress. We further evaluated the extent to which the setting of \emph{stagMax} determines the quality of results and convergence. Hence, for CGP TR with $\epsilon =$ 5\%, we tested the \emph{stagMax} equal to  1, 10, 50, and 100 generations. The resulting fitness progress is shown in Fig.~\ref{fig:comparison:stagnationparameter}.
The One-sided Mann–Whitney U test proves that the most suitable setting for \emph{stagMax} is 50 generations for the first 25\,s. After that, all the settings considered for \emph{stagMax} lead to results that are not  statistically different.

\subsection{Comparison with EvoApproxLib}
\label{sec:comp:evoapproxlib}

\begin{figure}[ht]
\centering
\includegraphics[width=0.99\columnwidth]{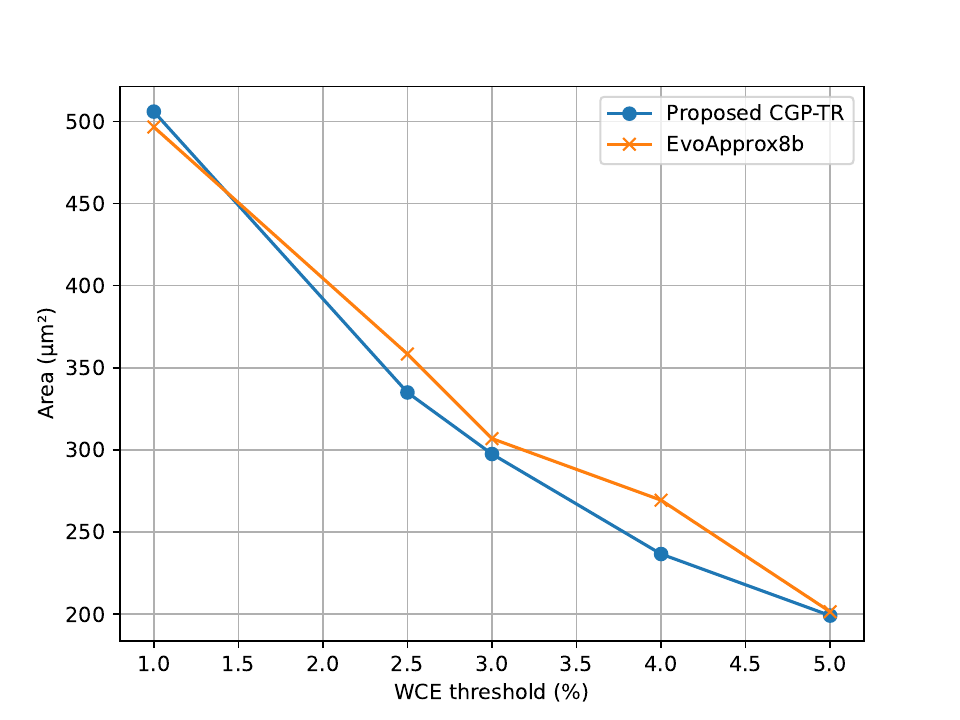}
\caption{WCE and area of the best approximate 8-bit multipliers from EvoApproxLib (orange color) and those evolved using CGP TR (blue color).}
\label{fig:comparison:bestmultipliers}
\end{figure}

The primary objective of our effort is to obtain approximate multipliers showing better trade-offs than the highly optimized multipliers from EvoApproxLib. Fig.~\ref{fig:comparison:bestmultipliers} shows properties of the best approximate multipliers according to WCE and area selected from 39,699 unsigned 8-bit approximate multipliers available in EvoApproxLib (orange color) and those evolved using CGP TR (blue color). 
In addition to the three WCE levels discussed in previous sections, the same experiments were performed for $\epsilon =$ 1\% and 2.5\%. Note that the area is always computed according to Table~\ref{tab:gates} for all multipliers considered in this study. With the exception of $\epsilon =$ 1\%, we concluded that the proposed method delivered approximate multipliers with improved properties. 

\section{Conclusions}
\label{sec:concl}

With the aim of producing approximate multipliers showing better properties than the highly optimized multipliers available in the EvoApproxLib, we introduced a transformer-based mutation to CGP. The transformer model guided CGP to more promising areas of the search space, especially when the standard mutation failed to escape from local optima. When the training time of the transformer model is not considered, approximate multipliers of the same quality are delivered faster by CGP TR than the standard CGP. Aside from the case where $\epsilon = 1\%$, we confirmed that the proposed method can generate approximate multipliers with improved characteristics. 

As an initial study on enhancing CGP with a transformer-based mutation operator, this work raises several directions for future research. In the present study, a separate transformer model was trained for each target WCE level, and experiments were limited to 8-bit unsigned approximate multipliers. An important open question is how to design and train a transformer-based mutation operator that generalizes across a broad range of WCE targets, varying bit widths, and potentially different classes of arithmetic circuits. Having such a universal transformer model for circuit approximation purposes would significantly improve design productivity.

This challenge is non-trivial because CGP exhibits highly specific genotype–phenotype–fitness relationships that are difficult to capture with statistical learning models. Another open issue concerns the construction of suitable training datasets for the transformer. For approximate circuits, CGP can be used to generate extensive datasets with desirable properties; however, it remains unclear which characteristics such datasets must possess to effectively support learning. Our future work will aim to address these questions.

\section*{Acknowledgements}
\vspace{3pt 
This work was supported by Czech Science Foundation project 24-10990S.
}

\bibliographystyle{IEEEtran}
\bibliography{IEEEabrv,wcci26_cgp_transf}

\begin{thebibliography}{10}
\providecommand{\url}[1]{#1}
\csname url@samestyle\endcsname
\providecommand{\newblock}{\relax}
\providecommand{\bibinfo}[2]{#2}
\providecommand{\BIBentrySTDinterwordspacing}{\spaceskip=0pt\relax}
\providecommand{\BIBentryALTinterwordstretchfactor}{4}
\providecommand{\BIBentryALTinterwordspacing}{\spaceskip=\fontdimen2\font plus
\BIBentryALTinterwordstretchfactor\fontdimen3\font minus \fontdimen4\font\relax}
\providecommand{\BIBforeignlanguage}[2]{{%
\expandafter\ifx\csname l@#1\endcsname\relax
\typeout{** WARNING: IEEEtran.bst: No hyphenation pattern has been}%
\typeout{** loaded for the language `#1'. Using the pattern for}%
\typeout{** the default language instead.}%
\else
\language=\csname l@#1\endcsname
\fi
#2}}
\providecommand{\BIBdecl}{\relax}
\BIBdecl

\bibitem{Mittal:2016}
S.~Mittal, ``A survey of techniques for approximate computing,'' \emph{ACM Comput. Surv.}, vol.~48, no.~4, pp. 62:1--62:33, 2016.

\bibitem{ArmeniakosZSH23}
G.~Armeniakos, G.~Zervakis, D.~Soudris, and J.~Henkel, ``Hardware approximate techniques for deep neural network accelerators: {A} survey,'' \emph{{ACM} Comput. Surv.}, vol.~55, no.~4, pp. 83:1--83:36, 2023.

\bibitem{Jiang:axc:surv:2020}
H.~Jiang, F.~J.~H. Santiago, H.~Mo, L.~Liu, and J.~Han, ``Approximate arithmetic circuits: {A} survey, characterization, and recent applications,'' \emph{Proc. {IEEE}}, vol. 108, no.~12, pp. 2108--2135, 2020.

\bibitem{miller:cgp:book}
J.~F. Miller, \emph{Cartesian Genetic Programming}, 1st~ed.\hskip 1em plus 0.5em minus 0.4em\relax Berlin, Germany: Springer-Verlag, 2011.

\bibitem{mrazek:date:17}
V.~Mrazek, R.~Hrbacek \emph{et~al.}, ``Evoapprox8b: Library of approximate adders and multipliers for circuit design and benchmarking of approximation methods,'' in \emph{Proc. of DATE'17}, 2017, pp. 258--261.

\bibitem{Mrazek:2020:approxMultipliersForCNN}
V.~Mrazek, L.~Sekanina, and Z.~Vasicek, ``Libraries of approximate circuits: Automated design and application in cnn accelerators,'' \emph{IEEE Journal on Emerging and Selected Topics in Circuits and Systems}, vol.~10, no.~4, pp. 406--418, 2020.

\bibitem{ceska:iccad17}
M.~Ceska, J.~Matyas, V.~Mrazek, L.~Sekanina, Z.~Vasicek, and T.~Vojnar, ``Approximating complex arithmetic circuits with formal error guarantees: 32-bit multipliers accomplished,'' in \emph{Proc. of 36th IEEE/ACM Int. Conf. On Computer Aided Design}.\hskip 1em plus 0.5em minus 0.4em\relax IEEE, 2017, pp. 416--423.

\bibitem{Ceska:sagtree:2022}
M.~Ceska, J.~Matys, V.~Mrazek, L.~Sekanina, Z.~Vasicek, and T.~Vojnar, ``Sagtree: Towards efficient mutation in evolutionary circuit approximation,'' \emph{Swarm Evol. Comput.}, vol.~69, p. 100986, 2022.

\bibitem{Wu:LLM:EA:2025}
X.~Wu, S.-H. Wu, J.~Wu, L.~Feng, and K.~C. Tan, ``Evolutionary computation in the era of large language model: Survey and roadmap,'' \emph{IEEE Transactions on Evolutionary Computation}, vol.~29, no.~2, pp. 534--554, 2025.

\bibitem{Hemberg2025}
E.~Hemberg, S.~Jorgensen, and U.-M. O'Reilly, \emph{Survey of Genetic Programming and Large Language Models}.\hskip 1em plus 0.5em minus 0.4em\relax Springer Nature Singapore, 2025, pp. 67--86.

\bibitem{Sobania:tec:2025}
D.~Sobania, J.~Petke, M.~Briesch, and F.~Rothlauf, ``A comparison of large language models and genetic programming for program synthesis,'' \emph{{IEEE} Trans. Evol. Comput.}, vol.~29, no.~4, pp. 1434--1448, 2025.

\bibitem{ACsurvey:ACM:2020}
P.~Stanley-Marbell, A.~Alaghi, M.~Carbin, E.~Darulova, L.~Dolecek, A.~Gerstlauer, G.~Gillani, D.~Jevdjic, T.~Moreau, M.~Cacciotti, A.~Daglis, N.~E. Jerger, B.~Falsafi, S.~Misailovic, A.~Sampson, and D.~Zufferey, ``Exploiting errors for efficiency: {A} survey from circuits to applications,'' \emph{ACM Comput. Surv.}, vol.~53, no.~3, 2020.

\bibitem{ALM}
\BIBentryALTinterwordspacing
W.~Liu, J.~Xu, D.~Wang, and F.~Lombardi, ``Design of approximate logarithmic multipliers,'' in \emph{Proceedings of the Great Lakes Symposium on VLSI 2017}, ser. GLSVLSI '17.\hskip 1em plus 0.5em minus 0.4em\relax New York, NY, USA: Association for Computing Machinery, 2017, p. 47–52. [Online]. Available: \url{https://doi.org/10.1145/3060403.3060409}
\BIBentrySTDinterwordspacing

\bibitem{Mahdiani:TCSI2009}
H.~R. Mahdiani, A.~Ahmadi, S.~M. Fakhraie, and C.~Lucas, ``Bio-inspired imprecise computational blocks for efficient vlsi implementation of soft-computing applications,'' \emph{IEEE Transactions on Circuits and Systems I: Regular Papers}, vol.~57, no.~4, pp. 850--862, April 2010.

\bibitem{WitschenMAP19}
L.~Witschen, H.~G. Mohammadi, M.~Artmann, and M.~Platzner, ``Jump search: {A} fast technique for the synthesis of approximate circuits,'' in \emph{Proceedings of the 2019 on Great Lakes Symposium on VLSI, {GLSVLSI}}.\hskip 1em plus 0.5em minus 0.4em\relax {ACM}, 2019, pp. 153--158.

\bibitem{Yi:GPTAC:2024}
S.~Yi, W.~Zuo, H.~Wu, R.~Dai, W.~Qian, and J.~Chen, ``Gptac: Domain-specific generative pre-trained model for approximate circuit design exploration,'' \emph{IEEE Journal on Emerging and Selected Topics in Circuits and Systems}, vol.~15, no.~2, pp. 349--360, 2025.

\bibitem{vaswani:2017:attentionneed}
A.~Vaswani, N.~Shazeer, N.~Parmar, J.~Uszkoreit, L.~Jones, A.~N. Gomez, {\L}.~Kaiser, and I.~Polosukhin, ``Attention is all you need,'' in \emph{Advances in Neural Information Processing Systems}, 2017, pp. 5998--6008.

\bibitem{devlin:2019:bert}
J.~Devlin, M.-W. Chang, K.~Lee, and K.~Toutanova, ``{BERT}: Pre-training of deep bidirectional transformers for language understanding,'' in \emph{Proc. of the 2019 Conf. of the North American Chapter of the Association for Computational Linguistics: Human Language Technologies, Volume 1}, vol.~1.\hskip 1em plus 0.5em minus 0.4em\relax Association for Computational Linguistics, Jun. 2019, pp. 4171--4186.

\bibitem{VeriGen:2024}
\BIBentryALTinterwordspacing
S.~Thakur, B.~Ahmad, H.~Pearce, B.~Tan, B.~Dolan{-}Gavitt, R.~Karri, and S.~Garg, ``Verigen: {A} large language model for verilog code generation,'' \emph{{ACM} Trans. Design Autom. Electr. Syst.}, vol.~29, no.~3, pp. 46:1--46:31, 2024. [Online]. Available: \url{https://doi.org/10.1145/3643681}
\BIBentrySTDinterwordspacing

\bibitem{EMLbook:2023}
W.~Banzhaf and P.~Machado, ``Fundamentals of evolutionary machine learning,'' in \emph{Handbook of Evolutionary Machine Learning}.\hskip 1em plus 0.5em minus 0.4em\relax Singapore: Springer Nature Singapore, 2024, pp. 3--28.

\bibitem{shem:2025:evoTrans}
\BIBentryALTinterwordspacing
E.~Shem-Tov, M.~Sipper, and A.~Elyasaf, ``Bert mutation: Deep transformer model for masked uniform mutation in genetic programming,'' \emph{Mathematics}, vol.~13, no.~5, 2025. [Online]. Available: \url{https://www.mdpi.com/2227-7390/13/5/779}
\BIBentrySTDinterwordspacing

\bibitem{CaetanoTP23}
V.~Caetano, M.~C. Teixeira, and G.~L. Pappa, ``Symbolic regression trees as embedded representations,'' in \emph{Proceedings of the Genetic and Evolutionary Computation Conference, {GECCO}}, S.~Silva and L.~Paquete, Eds.\hskip 1em plus 0.5em minus 0.4em\relax {ACM}, 2023, pp. 411--419.

\bibitem{TexPappa:gecco25}
M.~C. Teixeira and G.~L. Pappa, ``Transformers as surrogate models for genetic programming in automl tasks,'' in \emph{Proceedings of the Genetic and Evolutionary Computation Conference}, ser. GECCO '25.\hskip 1em plus 0.5em minus 0.4em\relax New York, NY, USA: ACM, 2025, p. 472–480.

\bibitem{Vasicek:2012:efficientCGP}
Z.~{Vašíček} and K.~{Slaný}, ``Efficient phenotype evaluation in cartesian genetic programming,'' in \emph{Proc. of the 15th European Conf. on Genetic Programming}, ser. LNCS, vol. 7244.\hskip 1em plus 0.5em minus 0.4em\relax Springer, 2012, pp. 266--278.

\bibitem{Hrbacek:2014:efficientCGP}
R.~{Hrbáček} and L.~{Sekanina}, ``Towards highly optimized cartesian genetic programming: From sequential via simd and thread to massive parallel implementation,'' in \emph{Proc. of Genetic and Evolutionary Computation}.\hskip 1em plus 0.5em minus 0.4em\relax ACM, 2014, pp. 1015--1022.

\bibitem{Pereyra:2017:Regularizing}
\BIBentryALTinterwordspacing
G.~Pereyra, G.~Tucker, J.~Chorowski, Łukasz Kaiser, and G.~Hinton, ``Regularizing neural networks by penalizing confident output distributions,'' 2017, arXiv preprint. [Online]. Available: \url{https://arxiv.org/abs/1701.06548}
\BIBentrySTDinterwordspacing

\end{thebibliography}

\end{document}